%% file: main.tex
\documentclass[conference]{IEEEtran}

\usepackage[T1]{fontenc}
\usepackage{mathptmx} 
\usepackage{cite}
\usepackage{amsmath,amssymb}
\usepackage{graphicx}
\usepackage{booktabs}
\usepackage{multirow}
\usepackage{xcolor}
\usepackage{url}

\title{Reliable Financial Named Entity Recognition under Domain Shift:\\
Confidence Estimation and Selective Prediction}

\author{
\IEEEauthorblockN{Zihao Zheng\IEEEauthorrefmark{1}, Baichuan Li\IEEEauthorrefmark{2},
Junyi Yao\IEEEauthorrefmark{1}, and Jiayu Long\IEEEauthorrefmark{1}}
\IEEEauthorblockA{\IEEEauthorrefmark{1}\textit{Washington University in St. Louis},
St. Louis, Missouri, USA\\
\IEEEauthorrefmark{2}\textit{Southern Methodist University}, Dallas, Texas, USA}
}

\begin{document}
\maketitle

\begin{abstract}
\input{sections/abstract}
\end{abstract}


\section{Introduction}
\input{sections/introduction}

\section{Related Work}
\input{sections/related_work}

\section{Data and Evaluation Protocol}
\input{sections/data}

\section{Confidence Signals}
\input{sections/method}

\section{Experimental Setup}
\input{sections/setup}

\section{Results}
\input{sections/results}

\section{Analysis and Deployment Implications}
\input{sections/analysis}

\section{Conclusion}
\input{sections/conclusion}

\section{Limitations and Ethical Considerations}
\input{sections/limitations}
\input{sections/ethics}

\bibliographystyle{IEEEtran}
\bibliography{references}

\end{document}

%% file: sections/abstract.tex
Financial AI systems often train information extractors on one textual
register and deploy them across filings, news, and user-generated content,
and standard F1 scores do not indicate which predictions remain safe to
automate when that input distribution changes. We study confidence
estimation and selective prediction for financial named entity recognition
(NER) on a three-tier stress test spanning SEC filings, financial news, and
general-topic social media as an extreme out-of-domain condition, evaluating
a BERT tagger and LoRA-tuned Qwen2.5-0.5B/1.5B models with five
inference-time confidence signals, three training seeds, and bootstrap
intervals. Confidence rankings themselves change under shift: whole-output
probability is the strongest in-domain error detector but deteriorates out of
domain, whereas entity-span probability and self-consistency are more robust;
self-consistency is also better calibrated without post-hoc fitting.
Abstention reduces sentence error from 34.3\% to below 2\% on the
highest-confidence 40\% of in-domain inputs and remains useful on financial
news, but recovers no usefully large clean subset under the extreme
social-media shift. These results motivate a staged deployment strategy
that detects severe distribution shift upstream before applying
prediction-level confidence gating.

%% file: sections/introduction.tex
Named entity recognition (NER) is a core information-extraction component
of applied financial AI systems for compliance, risk, and
market-intelligence workflows. A growing line of work fine-tunes small
open-weight language models for financial NER with parameter-efficient
methods and reports steadily rising F1 on in-domain test splits
\cite{lian2026llama,wu2026deepseek,lu2025financialner}. Yet purpose-built
financial LLMs still describe span-level NER as one of their hardest tasks
\cite{wu2023bloomberggpt,xie2023pixiu,xie2024finben}, and a deployed system
faces a question that leaderboard F1 does not answer: \emph{when the input
distribution drifts away from the training data, can the model tell us
which of its extractions to trust?}

This matters because financial text is not one domain: a model trained on
formal filings language (``\textit{Apple Inc.\ reported revenue of \$94.9
billion.}'') will meet news headlines and social-media commentary
(``\textit{AAPL just destroyed earnings}''). Prior work shows that neural
models become
\emph{overconfident} precisely when inputs shift away from training data
\cite{ovadia2019trust,kamath2020selective}, while complementary work detects
harmful covariate shift at the distribution level before unreliable
predictions reach downstream systems \cite{ginsberg2023harmful}. NER models
in particular lean on memorized entity surface forms rather than context
\cite{lin2021rockner}. In a financial setting, a hallucinated organization
or a mistyped person entity is not benign: such errors propagate into
downstream compliance, monitoring, and analytical pipelines.

We therefore study financial NER through the lens of \emph{calibrated
selective prediction} \cite{chow1970reject,elyaniv2010selective,
geifman2017selective}: allow the model to abstain on low-confidence
predictions and route them to human review, and ask how much reliability
this buys at what coverage cost. We fine-tune a BERT encoder tagger and
instruction-tuned generative models (Qwen2.5-0.5B and 1.5B-Instruct with
LoRA \cite{qwen25,hu2022lora}) on the FIN corpus of SEC filings
\cite{salinas2015fin} and evaluate them on a three-tier stress test:
in-domain filings (FIN test), financial news (FiNER-ORD
\cite{shah2023finerord}), and general-topic social media as an extreme
out-of-domain condition (TweetNER7 \cite{ushio2022tweetner7}). For the
generative models we compare five inference-time confidence signals that
require no additional training, spanning whole-output probabilities,
entity-restricted probabilities, and self-consistency vote share over
sampled decodes \cite{wang2023selfconsistency,manakul2023selfcheckgpt}.
Following \cite{kamath2020selective}, all thresholds and calibration
parameters are chosen on in-domain validation data and frozen before
touching any test set; results carry multi-seed variance and bootstrap
confidence intervals.

This study contributes a deployment-centered reliability evaluation for
financial NER: it demonstrates that confidence rankings are domain-sensitive
(the strongest in-domain signal can become inferior after a register or
topic shift), identifies entity-span probability and self-consistency as
complementary signals (stronger error ranking vs.\ better calibration), and
characterizes the operating boundary of confidence gating, whose
severe-shift failures motivate upstream domain-level screening as future
work.

%% file: sections/related_work.tex
Three lines of work underpin our study.

\paragraph{Financial NER}
The FIN dataset \cite{salinas2015fin} framed financial NER as domain
adaptation from newswire to SEC filings and remains the standard
filings-domain benchmark, adopted by FLUE \cite{shah2022flue}, BloombergGPT
\cite{wu2023bloomberggpt}, PIXIU \cite{xie2023pixiu}, and FinBen
\cite{xie2024finben}; FiNER-ORD \cite{shah2023finerord} provides manually
annotated financial \emph{news}. Recent work fine-tunes small LLMs for
financial NER with LoRA-style adapters \cite{lian2026llama,wu2026deepseek}
or evaluates prompting \cite{lu2025financialner}, reporting F1 on a
single-domain split without uncertainty analysis. We ask whether NER
confidence survives textual domain shift.

\paragraph{Selective prediction and calibration}
Abstention dates to Chow \cite{chow1970reject};
\cite{elyaniv2010selective,geifman2017selective} formalized the
risk--coverage trade-off, with maximum softmax probability (MSP) as the
canonical baseline \cite{hendrycks2017baseline}. Modern networks are
miscalibrated \cite{guo2017calibration}, and calibration degrades under
dataset shift \cite{ovadia2019trust,desai2020calibration}; complementary
work detects harmful covariate shift at the distribution level to identify
when deployment conditions may invalidate model generalization
\cite{ginsberg2023harmful}. Closest to our setting, Kamath et
al.~\cite{kamath2020selective} studied selective \emph{question answering}
under domain shift and showed softmax confidence becomes unreliable out of
domain; \cite{varshney2022investigating} cautioned that simple probability
baselines are hard to beat, which our results confirm in-domain and refute
under shift.

\paragraph{Uncertainty for NER and generative extraction}
Span-level confidence for extraction goes back to CRF-based estimates
\cite{culotta2004confidence}; later work calibrated entity-level confidences
of encoder taggers \cite{jagannatha2020calibrating}. Casting NER as text
generation \cite{wang2025gptner,zhou2024universalner,ding2024gner} makes
token probabilities, sampling agreement \cite{wang2023selfconsistency,
manakul2023selfcheckgpt}, and semantic entropy \cite{kuhn2023semantic,
farquhar2024hallucinations} available as confidence signals; sequence
probability requires length correction \cite{murray2018length}. Verbalized
confidence is systematically overconfident for small models
\cite{xiong2024canllms}, so we
restrict our study to logit- and sampling-based signals. To our knowledge,
prior financial NER work has not evaluated whether confidence rankings and
calibration survive such shifts; this is the gap we target. On the
deployment side, learning-to-defer \cite{mozannar2020defer} and
uncertainty-guided human--LLM work allocation \cite{li2023coannotating}
motivate our risk--coverage framing.

%% file: sections/data.tex
\label{sec:data}
We build a three-tier stress test from public datasets, harmonized to a
shared \{PER, ORG, LOC\} schema (Table~\ref{tab:datasets}). The first two
tiers stay within finance while changing register; the third intentionally
leaves the financial domain to probe a boundary condition under severe
distribution shift. Because the corpora also differ in annotation process,
topic, time, and entity density, the tiers are neither a controlled causal
decomposition of domain shift nor a purely gradual register change.

\paragraph{In-domain: FIN (SEC filings)}
The FIN corpus \cite{salinas2015fin} contains sentences from US SEC
financial agreements; its standard train/valid/test splits serve for
training, calibration/threshold fitting, and in-domain testing. Sentences
containing MISC entities are dropped (6 test sentences) so that all three
tiers share the same schema.

\paragraph{Near shift: FiNER-ORD (financial news)}
FiNER-ORD \cite{shah2023finerord} provides manually annotated PER/ORG/LOC
entities over financial news articles. The register shift from formal
filings boilerplate to journalistic prose constitutes our near-shift tier.

\paragraph{Far shift: TweetNER7 (social media)}
TweetNER7 \cite{ushio2022tweetner7} is a \emph{general-topic} Twitter NER
dataset, not a financial-tweet corpus. We use it only as an extreme
out-of-domain stress test for a model trained on financial filings. It
annotates seven entity types including \textit{corporation}, \textit{person},
and \textit{location}, which we map to ORG/PER/LOC. To keep gold labels
clean, our main far-shift set keeps only tweets whose entities all fall
inside the mapped schema, dropping the rest rather than silently relabeling.
This removes 2{,}179 of 2{,}807 tweets in the 2021 test split, leaving 628
eligible tweets (Table~\ref{tab:filtering}) --- a deliberate
precision-over-size choice quantified by a sensitivity analysis
(\S\ref{sec:analysis}) whose \emph{relaxed} variant keeps every tweet and
instead \emph{ignores} out-of-schema gold entities, so matching predictions
receive neither credit nor penalty. URL placeholders and user-mention markup
are normalized to plain surface forms. Both out-of-domain test sets are
capped at 300 sentences with a fixed subsampling seed.

\begin{table}[t]
\centering
\caption{Dataset statistics after harmonization and filtering.}
\label{tab:datasets}
\footnotesize
\setlength{\tabcolsep}{2pt}
\input{tables/datasets}
\vspace{2pt}

{\raggedright\footnotesize All filtering decisions are recorded in
machine-readable pipeline manifests.\par}
\end{table}

\subsection{Evaluation protocol}
\label{sec:protocol}
Encoder predictions are scored as exact BIO span matches. Generative
predictions are generated text, so we score them by \emph{position-less
multiset matching}: each predicted (surface form, type) pair is matched
against the multiset of gold pairs. Because the two model families are thus
scored under different matching procedures, direct encoder--generative F1
comparisons should be read as indicative rather than exact.
Surface forms are lowercased, whitespace-collapsed, and stripped of boundary
punctuation, so tokenization artifacts do not count as errors; duplicate
predictions can match at most as many gold occurrences as exist; wrong
boundaries or types count as both a false positive and a false negative.
Invalid JSON is salvaged with a tolerant parser and flagged; unparseable
output counts every gold entity as missed. Sentences without entities are
scored as correct exactly when the model predicts the empty set, with the
sequence probability as the confidence of that claim. Hallucinations ---
predicted surface forms absent from the source sentence --- are false
positives and additionally tracked as a separate rate.

%% file: tables/datasets.tex
\begin{tabular}{llrrrrr}
\toprule
Split & Domain & Sent. & Ent. & PER & ORG & LOC \\
\midrule
Train & FIN (filings) \cite{salinas2015fin} & 1014 & 980 & 648 & 175 & 157 \\
Valid & FIN (filings) \cite{salinas2015fin} & 150 & 177 & 97 & 68 & 12 \\
Test & FIN (filings) \cite{salinas2015fin} & 299 & 295 & 201 & 56 & 38 \\
Test & FiNER-ORD (news) \cite{shah2023finerord} & 300 & 322 & 78 & 151 & 93 \\
Test & TweetNER7 (tweets) \cite{ushio2022tweetner7} & 300 & 619 & 381 & 135 & 103 \\
\bottomrule
\end{tabular}

%% file: sections/method.tex
\label{sec:method}
\paragraph{Generative NER}
The generative model is instruction-tuned to map a sentence to a JSON object
$\{\texttt{"entities"}: [\{\texttt{"text"}, \texttt{"type"}\},\dots]\}$ and
scored under the protocol of \S\ref{sec:protocol}.

\paragraph{Confidence signals}
Let $y_{1:n}$ be the greedy output with token log-probabilities $\log
p(y_i)$. For each predicted entity $e$, we compare
\emph{sequence probability} (length-normalized)
$C_{\text{seq}} = \exp\big(\tfrac{1}{n}\sum_{i}\log p(y_i)\big)$
\cite{murray2018length}; \emph{token probability}
$C_{\text{tok}} = \tfrac{1}{n}\sum_{i} p(y_i)$; \emph{span probability}
$C_{\text{span}}(e) = \exp\big(\tfrac{1}{|T_e|}\sum_{i\in T_e}\log
p(y_i)\big)$,\newpage\vspace*{4pt} where $T_e$ are the tokens of $e$'s surface form; \emph{type
probability} $C_{\text{type}}(e)$, defined likewise over the tokens of $e$'s
type label; and \emph{self-consistency}
$C_{\text{sc}}(e) = \tfrac{1}{K}\sum_{k=1}^{K}\mathbf{1}[e \in \hat{E}_k]$,
the vote share of $e$ over $K$ sampled decodes $\hat{E}_k$.
$C_{\text{seq}}$ and $C_{\text{tok}}$ are sentence-level signals shared by
all entities in a sentence; the rest are entity-level. Self-consistency uses
$K{=}5$ samples at temperature $0.7$
\cite{wang2023selfconsistency,manakul2023selfcheckgpt}; an entity counts as
a vote when a sampled decode contains it (same normalized surface form and
type). A validation-selected convex combination of span and type confidence
selected pure span confidence (test-tier effect below $\pm$0.02 AUROC), so
we report span confidence directly (\S\ref{sec:analysis}).

\paragraph{Encoder baseline confidence}
For the BERT tagger, span confidence is the mean first-subword maximum
softmax probability within the predicted span
\cite{hendrycks2017baseline,culotta2004confidence}, optionally
temperature-scaled \cite{guo2017calibration} with $T$ fit on in-domain
validation token-level NLL.

\paragraph{Selective prediction}
We evaluate abstention at two granularities. \emph{Sentence level:} the
sentence confidence is the minimum entity confidence among its predictions
(sequence probability for empty predictions). Ranking sentences by this score
yields a risk--coverage curve \cite{elyaniv2010selective,geifman2017selective}:
risk is the fraction of answered sentences containing at least one extraction
error, coverage the fraction answered; we report its area (AURC).
\emph{Entity level:} following \cite{culotta2004confidence}, we rank all
emitted entities by confidence and report \emph{selective precision} at
fixed entity coverage. We additionally report entity-level error-detection
AUROC and expected calibration error (ECE, 15 equal-width bins)
\cite{naeini2015ece} against entity correctness; because binned ECE can be
unstable on test tiers of roughly 300 sentences, we interpret calibration
differences qualitatively and complement ECE with reliability diagrams
(\S\ref{sec:analysis}). Uncertainty is quantified as mean$\pm$std over
training seeds and 95\% bootstrap confidence intervals (1{,}000 resamples
over test instances, averaged over seeds).

%% file: sections/setup.tex
\label{sec:setup}
\paragraph{Models}
The encoder baseline is \texttt{bert-base-cased} with a token-classification
head (7 BIO labels), trained for 5 epochs (batch 16, lr $3\times10^{-5}$,
linear warmup 10\%). The primary generative model is
\texttt{Qwen2.5-0.5B-Instruct} \cite{qwen25} with LoRA \cite{hu2022lora}
adapters ($r{=}16$, $\alpha{=}32$, dropout 0.05, on q/k/v/o projections),
trained for 3 epochs (effective batch 16, lr $1\times10^{-4}$) with loss on
the JSON completion only, in fp16 with fp32 LoRA parameters. Both are trained
on the harmonized FIN training set with three random seeds (13/42/2026). To
test whether the confidence findings are an artifact of the smallest scale,
we additionally train \texttt{Qwen2.5-1.5B-Instruct} with the identical
recipe (single seed) as a scale check. All experiments ($\sim$25k
generations with log-probabilities and 5-sample consistency) ran on a single
consumer-grade machine (Apple M1 Pro, 16\,GB) in roughly two days;
self-consistency is the only signal with a multiplicative inference cost,
while span and sequence probabilities are free by-products of decoding.

\paragraph{Protocol}
The FIN validation split is the only data used for fitting: the encoder
temperature $T$ and any abstention thresholds are selected there and frozen
before evaluating any test domain \cite{kamath2020selective}; test sets are
never used for model or hyperparameter selection. Every pipeline stage
records the command line, input hashes, seeds, and library versions in
machine-readable manifests, and all figures and tables are generated
deterministically from per-entity and per-sentence prediction dumps.

%% file: sections/results.tex
\label{sec:results}

\subsection{Extraction quality degrades under shift}
Table~\ref{tab:main} gives the headline extraction results; the reported ECE
is entity-level, computed on span confidence for the generative models and on
MSP for the encoder. The encoder tagger drops from 70.3 micro-F1 in-domain to
38.3 on financial news and 24.1 on tweets --- a two-thirds relative loss.
The 0.5B generative model is substantially weaker in-domain (37.4$\pm$1.3
F1; encoder advantage $+$33.6 points, paired bootstrap 95\% CI
$[22.7, 44.5]$), reflecting its scale and the $\sim$1k-sentence training
budget, but degrades more gently: on the far-shift tier the encoder's
advantage vanishes (24.1$\pm$2.5 vs.\ 26.7$\pm$2.8; difference CI
$[-7.2, +2.0]$, consistent with parity), in line with reports that
instruction-tuned generative models generalize more gracefully
\cite{zhou2024universalner,ding2024gner}. Generation-specific failure modes
stay bounded but non-negligible: 1.4--5.4\% invalid JSON and 3.6--8.2\%
hallucinated entities across the three tiers.

\begin{table}[t]
\vspace*{1pt}
\centering
\caption{Main results (\%; mean$\pm$std over 3 seeds; Qwen2.5-1.5B is a
single-seed scale check).}
\label{tab:main}
\footnotesize
\setlength{\tabcolsep}{2.5pt}
\resizebox{\columnwidth}{!}{\input{tables/main_results}}
\end{table}

\subsection{Which confidence signals survive domain shift?}
Table~\ref{tab:signals} compares error-detection AUROC and calibration of
the five signals for the 0.5B model (mean$\pm$std over 3 seeds). Three
regularities emerge.

\textbf{(1) The best in-domain signal is not the best signal under shift.}
In-domain, whole-output probabilities are the strongest error detectors
(sequence 0.839$\pm$0.014), replicating the strength of simple probability
baselines \cite{hendrycks2017baseline,varshney2022investigating}. Under
shift their AUROC falls to 0.661/0.626 (news/tweets), while span-restricted
confidence holds at 0.690/0.721 and self-consistency at 0.671/0.682; on the
far tier the sequence and span bootstrap intervals separate ($[0.594,
0.656]$ vs.\ $[0.695, 0.748]$), so the reordering is not sampling noise.

\begin{figure}[t]
\centering
\includegraphics[width=0.86\linewidth]{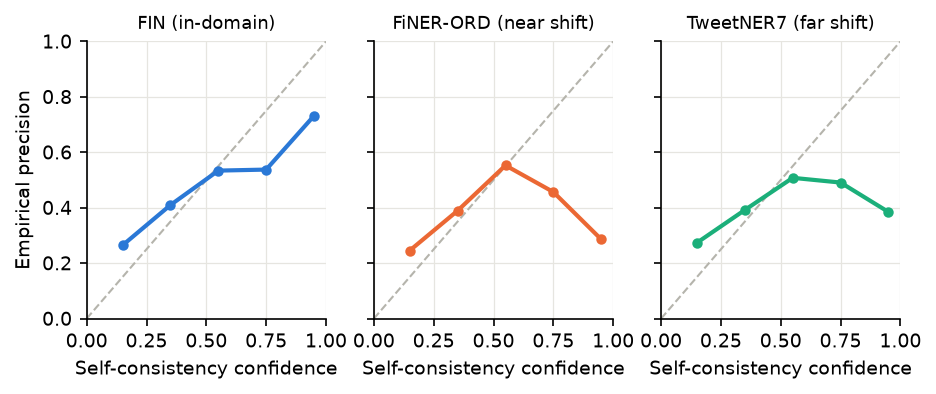}
\caption{Self-consistency reliability (0.5B, pooled seeds). The
highest-confidence bin inverts under both shifts.}
\label{fig:reliability}
\end{figure}

\textbf{(2) Discrimination and calibration dissociate.}
Probability-based signals are severely overconfident as probabilities
(span-signal entity-level ECE 0.36--0.41 across test tiers), while
self-consistency vote share is far better calibrated out of the box (ECE
0.10--0.12; Figure~\ref{fig:reliability}) --- its reliability curve tracks
the diagonal at low and middle confidence on every tier, but on both shifted
tiers the highest-confidence bin \emph{inverts}: under shift the model can
be consistently, confidently wrong, echoing \cite{ovadia2019trust}.

\textbf{(3) The encoder's confidence collapses under shift; the generative
model's does not.} The encoder's MSP is an excellent in-domain error
detector (AUROC 0.922$\pm$0.016) but drops to 0.633/0.615 under shift ---
the same failure Kamath et al.~\cite{kamath2020selective} report for
selective QA. The generative span signal loses far less
(0.801$\to$0.690/0.721), so under shift the \emph{weaker} model provides
the more useful uncertainty estimate.

\begin{table*}[t]
\centering
\caption{Error-detection AUROC and ECE per confidence signal
(Qwen2.5-0.5B; mean$\pm$std over 3 seeds), by test domain.}
\label{tab:signals}
\small
\input{tables/signal_comparison}
\end{table*}

\subsection{A single-seed scale check at 1.5B}
\label{sec:scale}
Table~\ref{tab:scale} repeats the signal comparison with Qwen2.5-1.5B
trained under the identical recipe but with a single seed, so the comparison
is indicative rather than a full replication. The larger model is
substantially stronger and more shift-robust (F1 46.2/52.1/47.7 across the
three tiers; far-tier sentence error rate 86\% vs.\ $\approx$95\% for the
0.5B model), and the study's two central regularities recur qualitatively.
First, whole-sequence probability again degrades most steeply as shift grows
(0.764$\to$0.637), while span probability matches or exceeds it on every
tier and self-consistency becomes the most shift-robust detector (0.744 on
news, 0.663 on tweets). Second, the calibration dissociation persists: span
ECE 0.23--0.36 vs.\ self-consistency ECE 0.11--0.17. On the far tier the
discrimination margins compress (all signals 0.64--0.66); as a single-seed
check, small differences here should not be over-read.

\begin{table}[t]
\centering
\caption{Scale check: error-detection AUROC by signal and domain for
Qwen2.5-0.5B (3-seed mean) vs.\ Qwen2.5-1.5B (single seed).}
\label{tab:scale}
\footnotesize
\setlength{\tabcolsep}{1pt}
\input{tables/scale_validation}
\end{table}

\subsection{Selective prediction: what abstention buys}
\begin{figure}[t]
\centering
\includegraphics[width=0.86\linewidth]{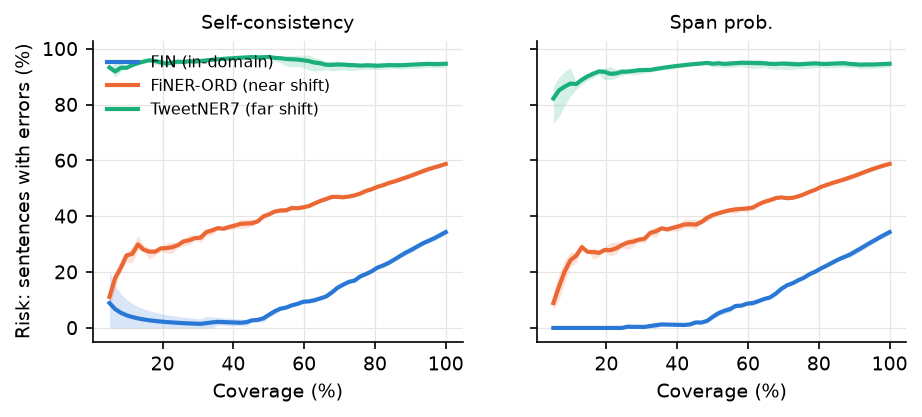}
\caption{Sentence-level risk--coverage for Qwen2.5-0.5B (3-seed mean and
range; lower is better).}
\label{fig:rc}
\end{figure}

Figure~\ref{fig:rc} shows sentence-level risk--coverage curves and
Table~\ref{tab:selective} entity-level selective precision. In-domain,
abstention is highly effective: answering the 40\% of sentences the span
signal trusts most drives sentence-level risk from 34.3\% to below 2\%
(every seed), and the encoder reaches 98.7\% entity precision at 50\%
coverage (from 66.7\% at full coverage). On the near-shift tier abstention
remains useful: risk falls from 58.8\% toward $\sim$15--25\% at low
coverage, and span-ranked selective precision rises $+$10 points at 60\%
coverage (30.7$\pm$1.3 $\to$ 40.8$\pm$2.6). On the far-shift tier, however,
$\approx$95\% of sentences contain an error and no signal finds a usefully
large clean subset (AURC 0.92--0.95): \emph{abstention complements, but
cannot substitute for, domain-appropriate training data}
(\S\ref{sec:analysis}).

\begin{table}[t]
\centering
\caption{Entity-level selective precision (\%) at 100/80/60\% coverage
(Qwen2.5-0.5B; mean$\pm$std over 3 seeds).}
\label{tab:selective}
\footnotesize
\setlength{\tabcolsep}{2pt}
\input{tables/selective_precision}
\end{table}

%% file: tables/main_results.tex
\begin{tabular}{llccccc}
\toprule
Model & Domain & P & R & F1 & Halluc.\% & ECE \\
\midrule
BERT-base & FIN & 66.7$\pm$2.3 & 74.2$\pm$1.4 & 70.3$\pm$1.5 & -- & 0.102$\pm$0.016 \\
BERT-base & FiNER-ORD & 39.1$\pm$0.3 & 37.6$\pm$2.3 & 38.3$\pm$1.1 & -- & 0.071$\pm$0.017 \\
BERT-base & TweetNER7 & 35.2$\pm$5.2 & 18.4$\pm$1.7 & 24.1$\pm$2.5 & -- & 0.074$\pm$0.007 \\
\midrule
Qwen-0.5B & FIN & 40.5$\pm$0.4 & 34.7$\pm$2.1 & 37.4$\pm$1.3 & 3.6$\pm$2.4 & 0.362$\pm$0.009 \\
Qwen-0.5B & FiNER-ORD & 30.7$\pm$1.3 & 18.0$\pm$4.4 & 22.6$\pm$3.8 & 8.2$\pm$0.7 & 0.407$\pm$0.005 \\
Qwen-0.5B & TweetNER7 & 30.8$\pm$0.7 & 23.7$\pm$4.2 & 26.7$\pm$2.8 & 5.6$\pm$1.1 & 0.370$\pm$0.007 \\
\midrule
Qwen-1.5B & FIN & 47.7 & 44.7 & 46.2 & 4.3 & 0.322 \\
Qwen-1.5B & FiNER-ORD & 56.3 & 48.4 & 52.1 & 3.6 & 0.227 \\
Qwen-1.5B & TweetNER7 & 44.9 & 50.9 & 47.7 & 6.8 & 0.358 \\
\bottomrule
\end{tabular}

%% file: tables/signal_comparison.tex
\begin{tabular}{lcccccc}
\toprule
Signal & \multicolumn{2}{c}{FIN} & \multicolumn{2}{c}{FiNER-ORD} & \multicolumn{2}{c}{TweetNER7} \\
 & AUROC & ECE & AUROC & ECE & AUROC & ECE \\
\midrule
Sequence prob. & 0.839$\pm$0.014 & 0.512$\pm$0.005 & 0.661$\pm$0.015 & 0.593$\pm$0.011 & 0.626$\pm$0.019 & 0.586$\pm$0.006 \\
Token prob. & 0.823$\pm$0.012 & 0.531$\pm$0.004 & 0.666$\pm$0.019 & 0.615$\pm$0.013 & 0.605$\pm$0.018 & 0.611$\pm$0.006 \\
Span prob. & 0.801$\pm$0.004 & 0.362$\pm$0.009 & 0.690$\pm$0.030 & 0.407$\pm$0.005 & 0.721$\pm$0.011 & 0.370$\pm$0.007 \\
Type prob. & 0.492$\pm$0.019 & 0.506$\pm$0.004 & 0.536$\pm$0.022 & 0.533$\pm$0.019 & 0.511$\pm$0.010 & 0.573$\pm$0.002 \\
Self-consistency & 0.690$\pm$0.019 & 0.116$\pm$0.029 & 0.671$\pm$0.048 & 0.120$\pm$0.015 & 0.682$\pm$0.017 & 0.100$\pm$0.008 \\
\bottomrule
\end{tabular}

%% file: tables/scale_validation.tex
\begin{tabular}{lcccccc}
\toprule
Signal & \multicolumn{3}{c}{0.5B (3 seeds)} & \multicolumn{3}{c}{1.5B (1 seed)} \\
 & FIN & FiNER & Tweet & FIN & FiNER & Tweet \\
\midrule
Sequence prob. & 0.839 & 0.661 & 0.626 & 0.764 & 0.720 & 0.637 \\
Span prob. & 0.801 & 0.690 & 0.721 & 0.819 & 0.729 & 0.638 \\
Self-consistency & 0.690 & 0.671 & 0.682 & 0.735 & 0.744 & 0.663 \\
\bottomrule
\end{tabular}

%% file: tables/selective_precision.tex
\begin{tabular}{llccc}
\toprule
Domain & Signal & P@100\% & P@80\% & P@60\% \\
\midrule
FIN & Seq. prob. & 40.5$\pm$0.4 & 49.0$\pm$0.8 & 58.4$\pm$0.6 \\
FIN & Span prob. & 40.5$\pm$0.4 & 49.2$\pm$0.3 & 56.4$\pm$0.6 \\
FIN & Self-consistency & 40.5$\pm$0.4 & 47.7$\pm$0.4 & 52.8$\pm$1.0 \\
\midrule
FiNER & Seq. prob. & 30.7$\pm$1.3 & 34.9$\pm$3.0 & 40.1$\pm$1.3 \\
FiNER & Span prob. & 30.7$\pm$1.3 & 35.3$\pm$1.7 & 40.8$\pm$2.6 \\
FiNER & Self-consist. & 30.7$\pm$1.3 & 34.9$\pm$3.0 & 41.4$\pm$2.3 \\
\midrule
Tweet & Seq. prob. & 30.8$\pm$0.7 & 34.2$\pm$1.0 & 36.8$\pm$2.9 \\
Tweet & Span prob. & 30.8$\pm$0.7 & 36.6$\pm$1.1 & 42.7$\pm$1.1 \\
Tweet & Self-consist. & 30.8$\pm$0.7 & 35.5$\pm$1.7 & 40.5$\pm$1.5 \\
\bottomrule
\end{tabular}

%% file: sections/analysis.tex
\label{sec:analysis}
\paragraph{Why is sequence probability vulnerable under shift while span
probability is more robust?}
The observed pattern is consistent with dilution by output scaffolding: once
the JSON format is learned, most generated tokens are structural and receive
probability near 1 regardless of input domain, so under shift this
structural floor compresses whole-output probability (mean sequence
probability stays above 0.9 on tweets even for wrong predictions).
Span-restricted confidence excludes the scaffolding. Consistently, the
validation-fit convex combination of span and type confidence
(\S\ref{sec:method}) selected pure span confidence in two of three seeds:
type confidence contributed no discriminative value (AUROC 0.49--0.54,
barely above chance). For these models, uncertainty lives in \emph{where the
entity is}, not \emph{what type it is}.

\paragraph{What do the two best signals disagree about?}
Span probability and self-consistency are complementary rather than
redundant: self-consistency provides substantially better calibration (ECE
$\le$0.12 vs.\ $\ge$0.36 without any post-hoc fitting), while span
confidence tends to provide the stronger error ranking under far shift
(AUROC 0.721 vs.\ 0.682, bootstrap intervals $[0.695,0.748]$ vs.\
$[0.653,0.710]$, overlapping only marginally) at $1\times$ rather than
$6\times$ decoding cost. A deployment can therefore choose by constraint:
span probability for a fixed-budget ranking gate, self-consistency when
thresholds must be expressed in probability units, or a trained calibrator
\cite{kamath2020selective} to combine both. We caution that setting a
threshold to a target precision (e.g., ``auto-accept above 80\% estimated
precision'') requires validating threshold-specific precision directly on
domain-matched data: binned ECE is estimated on $\approx$300-sentence tiers
where it can be unstable, and Figure~\ref{fig:reliability} shows aggregate
calibration can mask failures exactly in the high-confidence region such
thresholds rely on.

\paragraph{Where does abstention stop working?}
The general-topic far-shift tier exposes the boundary condition: with a
$\approx$95\% sentence-level base error rate there is no low-risk subset to
find, and the top-confidence self-consistency bin inverts
(Figure~\ref{fig:reliability}). Manual inspection shows the failure mode
RockNER \cite{lin2021rockner} predicts: the model consistently extracts
salient capitalized tokens as ORG/PER across all five samples ---
consistency measures stability of a bias, not correctness. This motivates a
staged deployment: (1) detect severe domain shift upstream, (2) apply
confidence-gated selective prediction within acceptable domains, (3) route
severe-shift inputs directly to human review. We do not implement or
evaluate such a domain detector here; distribution-level tests for harmful
covariate shift \cite{ginsberg2023harmful} are a natural candidate, and
building and benchmarking this routing stage is future work.

\paragraph{Sensitivity to TweetNER7 filtering}
\label{sec:sensitivity}
Table~\ref{tab:filtering} quantifies the far-tier filtering choice; the
generative comparison uses seed 42 under both protocols. The relaxed variant
leaves the picture unchanged: the generative model scores 29.1 F1 with span
AUROC 0.715 (vs.\ 29.7 / 0.725 on the main filtered set), and the encoder
drops slightly to 19.5$\pm$1.0 F1 (vs.\ 24.1$\pm$2.5) with statistically
indistinguishable MSP AUROC (0.650$\pm$0.047 vs.\ 0.615$\pm$0.054). If
anything, the main filtered set slightly \emph{understates} far-shift
difficulty; the headline conclusions are not artifacts of the filtering
rule.

\begin{table}[t]
\centering
\caption{TweetNER7 filtering sensitivity, main (drop-sentence) vs.\ relaxed
(ignore-entity) protocol. Encoder: 3-seed mean; Qwen: seed 42.}
\label{tab:filtering}
\footnotesize
\setlength{\tabcolsep}{2pt}
\input{tables/tweetner7_filtering}
\end{table}

\paragraph{Hallucination and abstention interact favorably}
Hallucinated entities (4--8\% of predictions) skew toward the low-confidence
tail: the bottom span-confidence tertile contains 67\% of hallucinated spans
on tweets and 61\% in-domain, versus 33\% under a uniform spread (39\% on
news). A confidence gate therefore preferentially removes fabricated
entities --- the highest-risk error class in a financial pipeline ---
exactly where fabrication is most frequent.

%% file: tables/tweetner7_filtering.tex
\begin{tabular}{lcc}
\toprule
 & Filtered & Relaxed \\
\midrule
Sentences dropped (of 2807) & 2179 & 0 \\
Evaluated (cap 300) & 300 & 300 \\
Gold PER/ORG/LOC & 619 & 443 \\
Ignored entities & 0 & 488 \\
\midrule
Encoder F1 & 24.1$\pm$2.5 & 19.5$\pm$1.0 \\
Encoder MSP AUROC & 0.615$\pm$0.054 & 0.650$\pm$0.047 \\
Qwen-0.5B F1 & 29.7 & 29.1 \\
Qwen-0.5B span AUROC & 0.725 & 0.715 \\
\bottomrule
\end{tabular}

%% file: sections/conclusion.tex
This study evaluated financial NER as a reliability component of applied AI
rather than only as an extraction benchmark. Confidence rankings change as
inputs move from filings to news and beyond the financial domain:
complete-output likelihood loses discrimination under shift, entity-span
probability is more robust, and self-consistency is better calibrated --- a
pattern that holds across three seeds and recurs qualitatively in a
single-seed 1.5B check. Selective prediction nearly eliminates error on the
highest-confidence in-domain subset and remains useful on financial news,
but cannot recover a usable clean subset when base extraction error is
extreme. A reliable financial AI pipeline should therefore screen for domain
shift upstream --- distribution-level shift tests \cite{ginsberg2023harmful}
are a natural building block --- apply confidence gating only within
supported domains, and route the rest to human review; implementing such a
routing stage is important future work.

%% file: sections/limitations.tex
\label{sec:limitations}
\textbf{Scale.} Our generative models (0.5B and 1.5B) and training set
($\sim$1k sentences) are small by production standards; absolute F1 values
are lower bounds for the setting. The 1.5B scale check rests on a single
seed, so its agreement with the 0.5B results is qualitative rather than a
demonstrated replication; whether the shift-degradation patterns attenuate
at much larger scale \cite{kadavath2022know} is open.

\textbf{Far-shift construction.} TweetNER7 is deliberately general-topic
rather than financial social media, and our main protocol retains 628 of
2{,}807 tweets (evaluating 300), so the far tier identifies an extreme
out-of-domain boundary but cannot isolate register shift or directly
represent financial social-media deployment; a matched, human-annotated
financial social-media test set would be the definitive fix. Our tiers also
vary jointly in register, topic, annotation, time period, and entity
density.

\textbf{Calibration measurement.} ECE is reported with 15 equal-width bins
on $\approx$300-sentence tiers; such estimates can be unstable and can
conceal high-confidence failures, which is why we pair them with reliability
diagrams. ECE confidence intervals and adaptive or classwise calibration
analyses are worthwhile in future, larger-scale studies.

\textbf{Signals and routing.} We exclude verbalized confidence (known to be
overconfident for small models \cite{xiong2024canllms}) and P(True)-style
self-verification \cite{kadavath2022know}, and semantic entropy
\cite{kuhn2023semantic} is only approximated by surface-form vote share.
Entity-level handoff, learned deferral \cite{mozannar2020defer}, and the
upstream domain detector our far-tier results motivate are not implemented
here and remain future work.

%% file: sections/ethics.tex
\textbf{Ethical considerations.} All datasets are publicly released research
corpora; we collect no new user data and do not attempt to identify
social-media users. Because confidence can remain high under shift,
consequential deployments require domain monitoring, human review, and
institution-specific validation; our results do not support automated
profiling or financial decisions about individuals.

%% file: main.bbl
\begin{thebibliography}{10}
\providecommand{\url}[1]{#1}
\csname url@samestyle\endcsname
\providecommand{\newblock}{\relax}
\providecommand{\bibinfo}[2]{#2}
\providecommand{\BIBentrySTDinterwordspacing}{\spaceskip=0pt\relax}
\providecommand{\BIBentryALTinterwordstretchfactor}{4}
\providecommand{\BIBentryALTinterwordspacing}{\spaceskip=\fontdimen2\font plus
\BIBentryALTinterwordstretchfactor\fontdimen3\font minus
  \fontdimen4\font\relax}
\providecommand{\BIBforeignlanguage}[2]{{%
\expandafter\ifx\csname l@#1\endcsname\relax
\typeout{** WARNING: IEEEtran.bst: No hyphenation pattern has been}%
\typeout{** loaded for the language `#1'. Using the pattern for}%
\typeout{** the default language instead.}%
\else
\language=\csname l@#1\endcsname
\fi
#2}}
\providecommand{\BIBdecl}{\relax}
\BIBdecl

\bibitem{lian2026llama}
Z.~Lian, ``Instruction finetuning llama-3-8b model using lora for financial
  named entity recognition,'' \emph{arXiv preprint arXiv:2601.10043}, 2026.

\bibitem{wu2026deepseek}
Y.~Wu and M.~Luo, ``Instruction finetuning {DeepSeek-R1-8B} model using {LoRA}
  and {NEFTune},'' \emph{arXiv preprint arXiv:2606.10392}, 2026.

\bibitem{lu2025financialner}
\BIBentryALTinterwordspacing
Y.-T. Lu and Y.~Huo, ``Financial named entity recognition: How far can llm
  go?'' in \emph{Proc. FinNLP--FNP--LLMFinLegal Workshop}, 2025, pp. 164--168.
  [Online]. Available: \url{https://aclanthology.org/2025.finnlp-1.15/}
\BIBentrySTDinterwordspacing

\bibitem{wu2023bloomberggpt}
S.~Wu \emph{et~al.}, ``Bloomberggpt: A large language model for finance,''
  \emph{arXiv preprint arXiv:2303.17564}, 2023.

\bibitem{xie2023pixiu}
Q.~Xie \emph{et~al.}, ``Pixiu: A comprehensive benchmark, instruction dataset
  and large language model for finance,'' in \emph{NeurIPS Datasets and
  Benchmarks}, 2023.

\bibitem{xie2024finben}
\BIBentryALTinterwordspacing
------, ``Finben: A holistic financial benchmark for large language models,''
  in \emph{NeurIPS Datasets and Benchmarks}, 2024. [Online]. Available:
  \url{https://proceedings.nips.cc/paper_files/paper/2024/hash/adb1d9fa8be4576d28703b396b82ba1b-Abstract-Datasets_and_Benchmarks_Track.html}
\BIBentrySTDinterwordspacing

\bibitem{ovadia2019trust}
Y.~Ovadia \emph{et~al.}, ``Can you trust your model's uncertainty? evaluating
  predictive uncertainty under dataset shift,'' in \emph{Proc. NeurIPS}, 2019.

\bibitem{kamath2020selective}
\BIBentryALTinterwordspacing
A.~Kamath, R.~Jia, and P.~Liang, ``Selective question answering under domain
  shift,'' in \emph{Proc. ACL}, 2020. [Online]. Available:
  \url{https://aclanthology.org/2020.acl-main.503/}
\BIBentrySTDinterwordspacing

\bibitem{ginsberg2023harmful}
\BIBentryALTinterwordspacing
T.~Ginsberg, Z.~Liang, and R.~G. Krishnan, ``A learning based hypothesis test
  for harmful covariate shift,'' in \emph{Proc. ICLR}, 2023. [Online].
  Available: \url{https://openreview.net/forum?id=rdfgqiwz7lZ}
\BIBentrySTDinterwordspacing

\bibitem{lin2021rockner}
B.~Y. Lin, W.~Gao, J.~Yan, R.~Moreno, and X.~Ren, ``Rockner: A simple method to
  create adversarial examples for evaluating the robustness of named entity
  recognition models,'' in \emph{Proc. EMNLP}, 2021.

\bibitem{chow1970reject}
C.~K. Chow, ``On optimum recognition error and reject tradeoff,'' \emph{IEEE
  Trans. Inf. Theory}, vol.~16, no.~1, pp. 41--46, 1970.

\bibitem{elyaniv2010selective}
\BIBentryALTinterwordspacing
R.~El-Yaniv and Y.~Wiener, ``On the foundations of noise-free selective
  classification,'' \emph{J. Mach. Learn. Res.}, vol.~11, pp. 1605--1641, 2010.
  [Online]. Available: \url{https://jmlr.org/papers/v11/el-yaniv10a.html}
\BIBentrySTDinterwordspacing

\bibitem{geifman2017selective}
\BIBentryALTinterwordspacing
Y.~Geifman and R.~El-Yaniv, ``Selective classification for deep neural
  networks,'' in \emph{Proc. NeurIPS}, 2017. [Online]. Available:
  \url{https://proceedings.neurips.cc/paper/2017/hash/4a8423d5e91fda00bb7e46540e2b0cf1-Abstract.html}
\BIBentrySTDinterwordspacing

\bibitem{qwen25}
{Qwen Team}, ``Qwen2.5 technical report,'' \emph{arXiv preprint
  arXiv:2412.15115}, 2024.

\bibitem{hu2022lora}
E.~J. Hu \emph{et~al.}, ``Lora: Low-rank adaptation of large language models,''
  in \emph{Proc. ICLR}, 2022.

\bibitem{salinas2015fin}
\BIBentryALTinterwordspacing
J.~C. Salinas~Alvarado, K.~Verspoor, and T.~Baldwin, ``Domain adaption of named
  entity recognition to support credit risk assessment,'' in \emph{Proceedings
  of the Australasian Language Technology Association Workshop 2015}, 2015.
  [Online]. Available: \url{https://aclanthology.org/U15-1010/}
\BIBentrySTDinterwordspacing

\bibitem{shah2023finerord}
A.~Shah, A.~Gullapalli, R.~Vithani, M.~Galarnyk, and S.~Chava, ``Finer-ord:
  Financial named entity recognition open research dataset,'' \emph{arXiv
  preprint arXiv:2302.11157}, 2023.

\bibitem{ushio2022tweetner7}
A.~Ushio, L.~Neves, V.~Silva, F.~Barbieri, and J.~Camacho-Collados, ``Named
  entity recognition in twitter: A dataset and analysis on short-term temporal
  shifts,'' in \emph{Proc. AACL-IJCNLP}, 2022.

\bibitem{wang2023selfconsistency}
X.~Wang \emph{et~al.}, ``Self-consistency improves chain of thought reasoning
  in language models,'' in \emph{Proc. ICLR}, 2023.

\bibitem{manakul2023selfcheckgpt}
P.~Manakul, A.~Liusie, and M.~J.~F. Gales, ``Selfcheckgpt: Zero-resource
  black-box hallucination detection for generative large language models,'' in
  \emph{Proc. EMNLP}, 2023.

\bibitem{shah2022flue}
\BIBentryALTinterwordspacing
R.~Shah \emph{et~al.}, ``When flue meets flang: Benchmarks and large pretrained
  language model for financial domain,'' in \emph{Proc. EMNLP}, 2022. [Online].
  Available: \url{https://aclanthology.org/2022.emnlp-main.148/}
\BIBentrySTDinterwordspacing

\bibitem{hendrycks2017baseline}
D.~Hendrycks and K.~Gimpel, ``A baseline for detecting misclassified and
  out-of-distribution examples in neural networks,'' in \emph{Proc. ICLR},
  2017.

\bibitem{guo2017calibration}
\BIBentryALTinterwordspacing
C.~Guo, G.~Pleiss, Y.~Sun, and K.~Q. Weinberger, ``On calibration of modern
  neural networks,'' in \emph{Proc. ICML}, 2017. [Online]. Available:
  \url{https://proceedings.mlr.press/v70/guo17a.html}
\BIBentrySTDinterwordspacing

\bibitem{desai2020calibration}
\BIBentryALTinterwordspacing
S.~Desai and G.~Durrett, ``Calibration of pre-trained transformers,'' in
  \emph{Proc. EMNLP}, 2020. [Online]. Available:
  \url{https://aclanthology.org/2020.emnlp-main.21/}
\BIBentrySTDinterwordspacing

\bibitem{varshney2022investigating}
\BIBentryALTinterwordspacing
N.~Varshney, S.~Mishra, and C.~Baral, ``Investigating selective prediction
  approaches across several tasks in iid, ood, and adversarial settings,'' in
  \emph{Findings of ACL}, 2022. [Online]. Available:
  \url{https://aclanthology.org/2022.findings-acl.158/}
\BIBentrySTDinterwordspacing

\bibitem{culotta2004confidence}
\BIBentryALTinterwordspacing
A.~Culotta and A.~McCallum, ``Confidence estimation for information
  extraction,'' in \emph{Proc. HLT-NAACL}, 2004. [Online]. Available:
  \url{https://aclanthology.org/N04-4028/}
\BIBentrySTDinterwordspacing

\bibitem{jagannatha2020calibrating}
\BIBentryALTinterwordspacing
A.~Jagannatha and H.~Yu, ``Calibrating structured output predictors for natural
  language processing,'' in \emph{Proc. ACL}, 2020. [Online]. Available:
  \url{https://aclanthology.org/2020.acl-main.188/}
\BIBentrySTDinterwordspacing

\bibitem{wang2025gptner}
\BIBentryALTinterwordspacing
S.~Wang \emph{et~al.}, ``Gpt-ner: Named entity recognition via large language
  models,'' in \emph{Findings of NAACL}, 2025. [Online]. Available:
  \url{https://aclanthology.org/2025.findings-naacl.239/}
\BIBentrySTDinterwordspacing

\bibitem{zhou2024universalner}
W.~Zhou, S.~Zhang, Y.~Gu, M.~Chen, and H.~Poon, ``Universalner: Targeted
  distillation from large language models for open named entity recognition,''
  in \emph{Proc. ICLR}, 2024.

\bibitem{ding2024gner}
\BIBentryALTinterwordspacing
Y.~Ding, J.~Li, P.~Wang, Z.~Tang, B.~Yan, and M.~Zhang, ``Rethinking negative
  instances for generative named entity recognition,'' in \emph{Findings of
  ACL}, 2024. [Online]. Available:
  \url{https://aclanthology.org/2024.findings-acl.206/}
\BIBentrySTDinterwordspacing

\bibitem{kuhn2023semantic}
L.~Kuhn, Y.~Gal, and S.~Farquhar, ``Semantic uncertainty: Linguistic
  invariances for uncertainty estimation in natural language generation,'' in
  \emph{Proc. ICLR}, 2023.

\bibitem{farquhar2024hallucinations}
S.~Farquhar, J.~Kossen, L.~Kuhn, and Y.~Gal, ``Detecting hallucinations in
  large language models using semantic entropy,'' \emph{Nature}, vol. 630, pp.
  625--630, 2024.

\bibitem{murray2018length}
\BIBentryALTinterwordspacing
K.~Murray and D.~Chiang, ``Correcting length bias in neural machine
  translation,'' in \emph{Proc. WMT}, 2018. [Online]. Available:
  \url{https://aclanthology.org/W18-6322/}
\BIBentrySTDinterwordspacing

\bibitem{xiong2024canllms}
M.~Xiong \emph{et~al.}, ``Can llms express their uncertainty? an empirical
  evaluation of confidence elicitation in llms,'' in \emph{Proc. ICLR}, 2024.

\bibitem{mozannar2020defer}
\BIBentryALTinterwordspacing
H.~Mozannar and D.~Sontag, ``Consistent estimators for learning to defer to an
  expert,'' in \emph{Proc. ICML}, 2020. [Online]. Available:
  \url{https://proceedings.mlr.press/v119/mozannar20b.html}
\BIBentrySTDinterwordspacing

\bibitem{li2023coannotating}
M.~Li \emph{et~al.}, ``Coannotating: Uncertainty-guided work allocation between
  human and large language models for data annotation,'' in \emph{Proc. EMNLP},
  2023.

\bibitem{naeini2015ece}
M.~Pakdaman~Naeini, G.~F. Cooper, and M.~Hauskrecht, ``Obtaining well
  calibrated probabilities using bayesian binning,'' in \emph{Proc. AAAI},
  2015.

\bibitem{kadavath2022know}
S.~Kadavath \emph{et~al.}, ``Language models (mostly) know what they know,''
  \emph{arXiv preprint arXiv:2207.05221}, 2022.

\end{thebibliography}
